\documentclass[a4paper,fleqn]{cas-sc}

\usepackage[authoryear,longnamesfirst]{natbib}
\usepackage{natbib}
\setcitestyle{numbers,square}
\def\tsc#1{\csdef{#1}{\textsc{\lowercase{#1}}\xspace}}
\tsc{WGM}
\tsc{QE}

\begin{document}
\let\WriteBookmarks\relax
\def\floatpagepagefraction{1}
\def\textpagefraction{.001}

\shorttitle{Summary of ChatGPT-Related Research}    

\shortauthors{Yiheng Liu et al.}  

\title [mode = title]{Summary of ChatGPT-Related Research and Perspective Towards the Future of Large Language Models}  



%

\author[1]{Yiheng Liu $^*$\corref{co}}
\author[1]{Tianle Han $^*$\corref{co}}
\author[1]{Siyuan Ma}
\author[1]{Jiayue Zhang}
\author[1]{Yuanyuan Yang}
\author[1]{Jiaming Tian}
\author[1]{Hao He}
\author[2]{Antong Li}
\author[1]{Mengshen He}
\author[3]{Zhengliang Liu}
\author[3]{Zihao Wu}
\author[3]{Lin Zhao}
\author[4]{Dajiang Zhu}
\author[5]{Xiang Li}
\author[1]{Ning Qiang}
\author[6,7,8]{Dingang Shen}
\author[3]{Tianming Liu}
\author[1]{Bao Ge \corref{cor1}}

\cortext[co]{These authors contributed equally to this work.}

\cortext[cor1]{Corresponding author}









\affiliation[1]{organization={School of Physics and Information Technology, Shaanxi Normal University},
        city={Xi'an},
        postcode={710119}, 
        state={Shaanxi},
        country={China}}
    
\affiliation[2]{organization={School of Life and Technology 
	Biomedical-Engineering, Xi'an Jiaotong University},
       	city={Xi'an},
       	postcode={710119}, 
       	state={Shaanxi},
       	country={China}}

\affiliation[3]{organization={School of Computing, The University of Georgia},
city={Athens},
postcode={30602}, 
country={USA}}

\affiliation[4]{organization={Department of Computer Science and Engineering, The University of Texas at Arlington},
city={Arlington},
postcode={76019}, 
country={USA}}

\affiliation[5]{organization={Department of Radiology, Massachusetts General Hospital and Harvard Medical School},
city={Boston},
postcode={02115}, 
country={USA}}

\affiliation[6]{organization={School of Biomedical Engineering, ShanghaiTech University},
city={Shanghai},
postcode={201210}, 
country={China}}

\affiliation[7]{organization={Shanghai United Imaging Intelligence Co., Ltd.},
city={Shanghai},
postcode={200230}, 
country={China}}

\affiliation[8]{organization={Shanghai Clinical Research and Trial Center},
city={Shanghai},
postcode={201210}, 
country={China}}




\begin{abstract}
This paper presents a comprehensive survey of ChatGPT-related (GPT-3.5 and GPT-4) research, state-of-the-art large language models (LLM) from the GPT series, and their prospective applications across diverse domains. Indeed, key innovations such as large-scale pre-training that captures knowledge across the entire world wide web, instruction fine-tuning and Reinforcement Learning from Human Feedback (RLHF) have played significant roles in enhancing LLMs' adaptability and performance. We performed an in-depth analysis of 194 relevant papers on arXiv, encompassing trend analysis, word cloud representation, and distribution analysis across various application domains. The findings reveal a significant and increasing interest in ChatGPT-related research, predominantly centered on direct natural language processing applications, while also demonstrating considerable potential in areas ranging from education and history to mathematics, medicine, and physics. This study endeavors to furnish insights into ChatGPT's capabilities, potential implications, ethical concerns, and offer direction for future advancements in this field.
\end{abstract}


\begin{highlights}
\item A comprehensive survey of ChatGPT-related research.
\item Analysis of 194 research papers.
\item Paving the way for further research and exploration in leveraging large language models for various applications.
\end{highlights}
\maketitle

\section{Introduction}
Recent advances in natural language processing (NLP) have led to the development of powerful language models such as the GPT (Generative Pre-trained Transformer) series \cite{GPT,GPT2,GPT-2,GPT3,instructGPT}, including large language models (LLM) such as ChatGPT (GPT-3.5 and GPT-4) \cite{openai2023gpt4}. These models are pre-trained on vast amounts of text data and have demonstrated exceptional performance in a wide range of NLP tasks, including language translation, text summarization, and question-answering. In particular, the ChatGPT model has demonstrated its potential in various fields, including education, healthcare, reasoning, text generation, human-machine interaction, and scientific research.

A key milestone of LLM development is InstructGPT \cite{instructGPT}, a framework that allows for instruction fine-tuning of a pre-trained language model based on Reinforcement Learning from Human Feedback (RLHF) \cite{RLHF,instructGPT}. This framework enables an LLM to adapt to a wide range of NLP tasks, making it highly versatile and flexible by leveraging human feedback. RLHF enables the model to align with human preferences and human values, which significantly improves from large language models that are solely trained text corpora through unsupervised pre-training. ChatGPT is a successor to InstructGPT. Since its release in December 2022, ChatGPT has been equipped with these advanced developments, leading to impressive performances in various downstream NLP tasks such as reasoning and generalized text generation. These unprecedented NLP capabilities spur applications in diverse domains such as education, healthcare, human-machine interaction, medicine and scientific research. ChatGPT has received widespread attention and interest, leading to an increasing number of applications and research that harness its exceeding potential. The open release of the multi-modal GPT-4 model further expands the horizon of large language models and empowers exciting developments that involve diverse data beyond text. 

\begin{figure}
	\centerline{\includegraphics[width=\columnwidth]{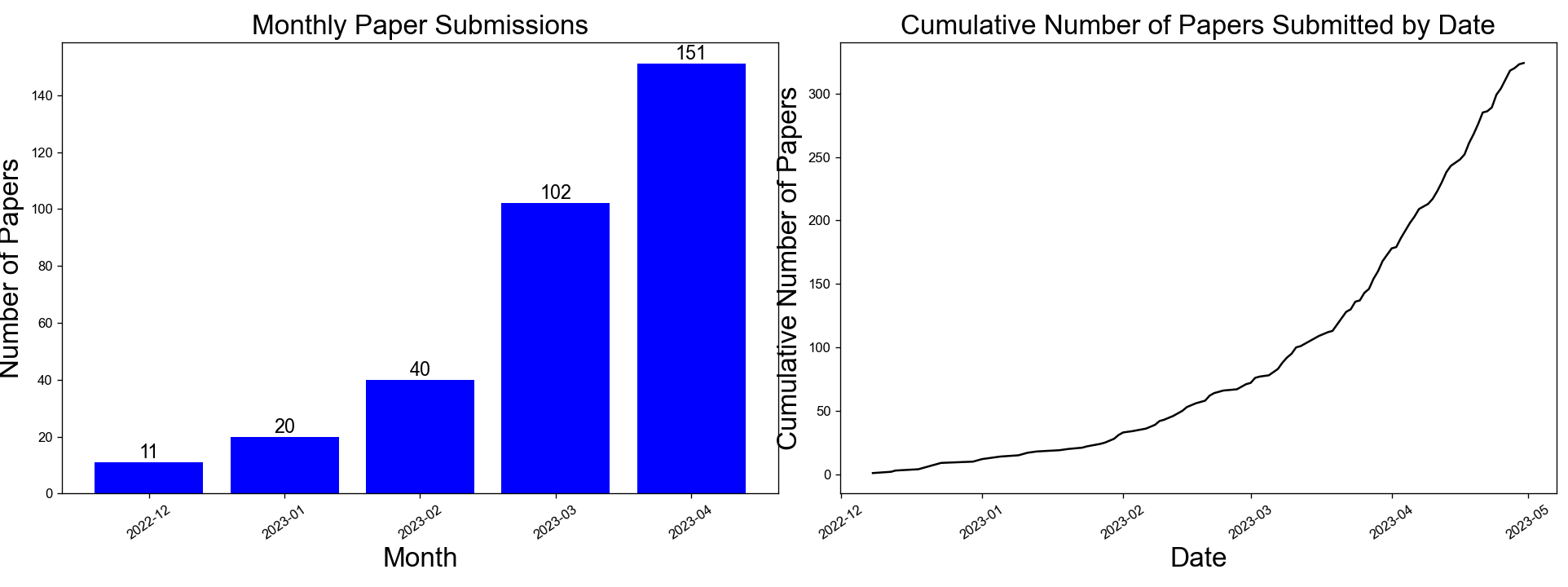}}
	\caption{The graphical representation is utilized to depict the number of research articles related to ChatGPT published from 2022 to April, 2023, revealing the trend and growth of ChatGPT-related research over time. The graph showcases the monthly count of submissions and cumulative daily submitted count in arXiv. Over time, there has been an increasing amount of research related to ChatGPT.}
	\label{fig:date}
\end{figure}

\begin{figure}
	\centerline{\includegraphics[width=\columnwidth]{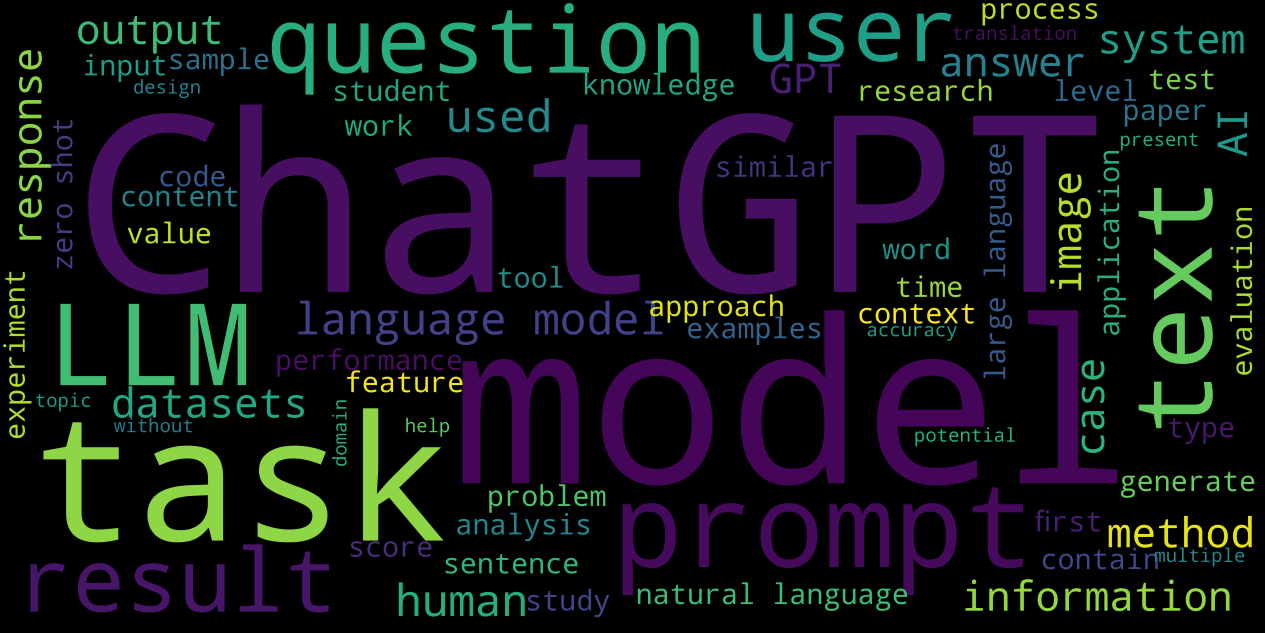}}
	\caption{Word cloud analysis of all the 194 papers.}
	\label{fig:wordcloud}
\end{figure}

\begin{figure}
	\centerline{\includegraphics[width=\columnwidth]{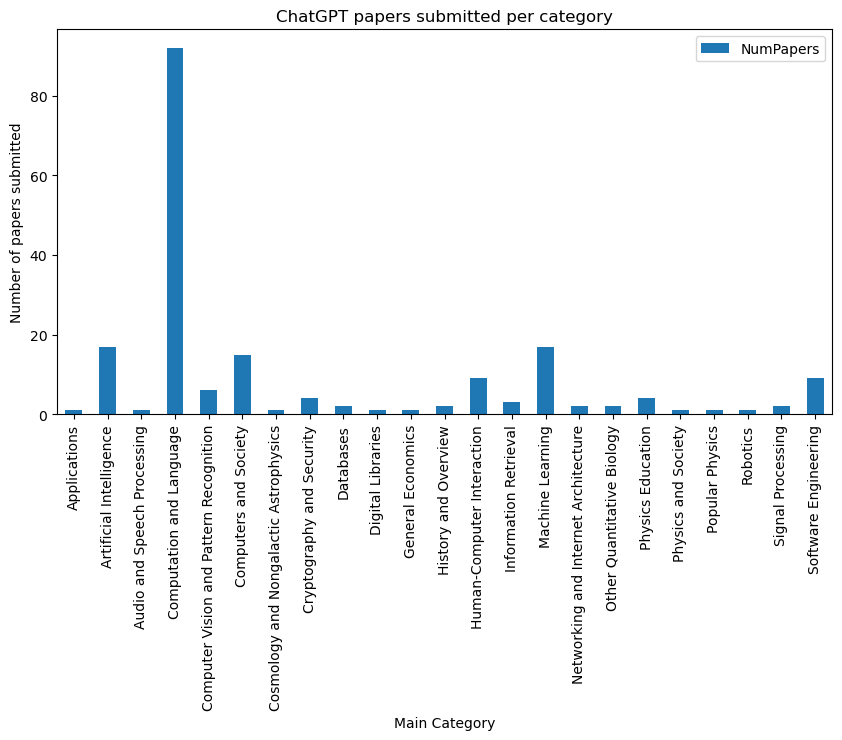}}
	\caption{The distribution of submitted papers across various fields.}
	\label{fig:category}
\end{figure}

The purpose of this paper is to provide a comprehensive survey of the existing research on ChatGPT and its potential applications in various fields. To achieve this goal, we conducted a thorough analysis of papers related to ChatGPT in the arXiv repository. As of April 1st, 2023, there are a total of 194 papers mentioning ChatGPT on arXiv. In this study, we conducted a trend analysis of these papers and generated a word cloud to visualize the commonly used terms. Additionally, we also examined the distribution of the papers across various fields and presented the corresponding statistics. Figure~\ref{fig:date} displays the submission trend of papers related to ChatGPT, indicating a growing interest in this field. Figure~\ref{fig:wordcloud} illustrates the word cloud analysis of all the papers. We can observe that the current research is primarily focused on natural language processing, but there is still significant potential for research in other fields such as education, medical and history. This is further supported by Figure~\ref{fig:category}, which displays the distribution of submitted papers across various fields, highlighting the need for more research and development in these areas. Due to the rapid advancement in research related to ChatGPT, we have also introduced a dynamic webpage that provides real-time updates on the latest trends in these area. Interested readers can access the webpage and stay informed about the evolving research directions by following this link \footnote{\url{https://snnubiai.github.io/chatgpt_arxiv_analysis/}}.

\begin{figure}
	\centerline{\includegraphics[width=\columnwidth]{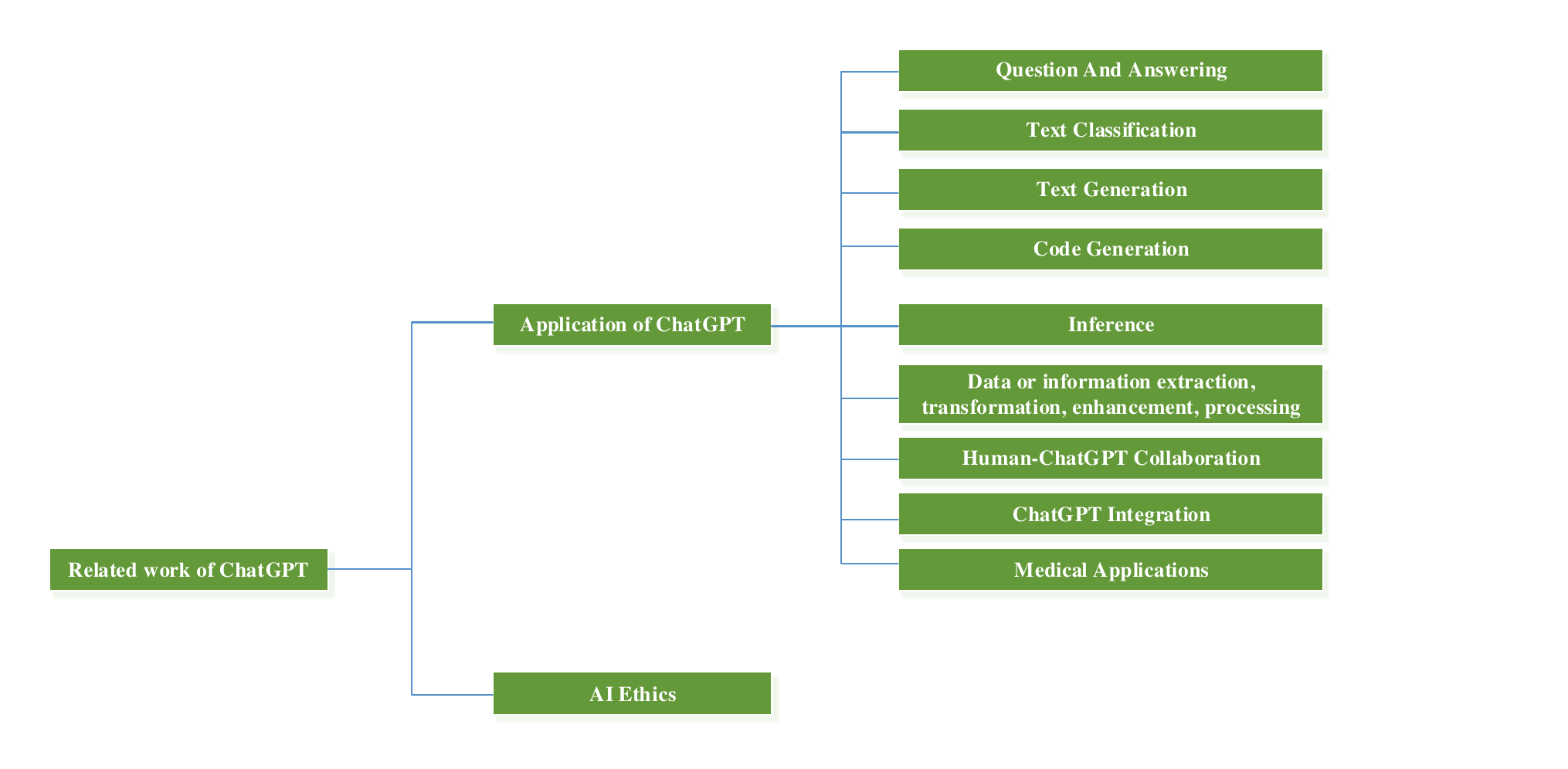}}
	\caption{Structure Diagram of Chapter 2.}
	\label{fig:structure}
\end{figure}

This paper aims to shed light on the promising capabilities of ChatGPT and provide insight into its potential impact in the future, including ethical considerations. Through this survey, we hope to provide insights into how these models can be improved and extended in the future. In section~\ref{sec:related}, we will review the existing work related to ChatGPT, including its applications and ethical considerations. In section~\ref{sec::evaluate}, we conducted a review of existing literature that assesses the capabilities of ChatGPT. We comprehensively evaluated the performance of ChatGPT based on these studies. In addition to discussing the current state of research related to ChatGPT, we will also explore its limitations in section~\ref{sec:future}. Furthermore, we will provide guidance on future directions for language model development.

\section{Related work of ChatGPT}
\label{sec:related}
In this section, we review the latest research related to the application and ethics of ChatGPT. Figure~\ref{fig:structure} shows the overall framework of this part. 

\subsection{Application of ChatGPT}
\subsubsection{Question And Answering}
\vspace{5mm}
\textbf{In the field of education}

ChatGPT is commonly used for question and answers testing in the education sector. Users can use ChatGPT to learn, compare and verify answers for different academic subjects such as physics, mathematics, and chemistry, and/or conceptual subjects such as philosophy and religion. Additionally, users can ask open-ended and analytical questions to understand the capabilities of ChatGPT. 

In the field of mathematics, Frieder et al. \cite{40} constructed the GHOSTS natural language dataset, which consists of graduate-level math test questions. The authors tested ChatGPT's math abilities on the GHOSTS dataset using a question-and-answer format and evaluated it according to fine-grained standards.In the Grad Text dataset, which covers simple set theory and logic problems, ChatGPT performed the best. However, in the Olympiad-Problem-Solving dataset, ChatGPT performed poorly, receiving only two 4-point scores (out of a total of 5), with the majority of scores being 2 points. In the Holes-in-Proofs dataset, ChatGPT received the lowest score of 1 point. In the MATH dataset, ChatGPT only scored impressively in 26\% of cases. These results suggest that ChatGPT's math abilities are clearly lower than those of ordinary math graduate students. Although ChatGPT can generally understand math problems, it fails to provide the correct solutions. Pardos et al. \cite{57} used the Open Adaptive Tutoring system (OATutor) to investigate whether prompts generated by ChatGPT were helpful for learning algebra, with 77 participants from Mechanical Turk taking part in the experiment. The experiment used questions from OpenStax's Elementary and Intermediate Algebra textbooks. These participants were randomly assigned to either a control group (with manual prompts) or an experimental group (with ChatGPT prompts). For each question in both courses, the authors obtained answers from ChatGPT through a question-and-answer format and evaluated scores according to three criteria: ChatGPT provided an answer, the answer was correct, and inappropriate language was not used in the answer. The study found that 70\% of prompts generated by ChatGPT passed manual quality checks, and both humans and ChatGPT produced positive learning gains. However, the scores of human prompts ranged from 74.59\% to 84.32\%, significantly higher than those of ChatGPT prompts. Shakarian et al. \cite{83} studied the performance of ChatGPT on math word problems (MWPs), using the DRAW-1K dataset for experimentation. The dataset consists of 1000 MWPs and their answers, along with algebraic equation templates for solving such problems. The authors used the idea of machine learning introspection and built performance prediction models using random forests and XGBoost, and evaluated them on the dataset using five-fold cross-validation. ChatGPT's accuracy increased from an initial 34\% to a final 69\%, while its recall increased from an initial 41\% to a final 83\%. The authors also found that ChatGPT's failure rate decreased from an initial 84\% to a final 20\%, indicating that performance can vary greatly depending on specific job requirements.

In the field of physics, Lehnert et al. \cite{27} explored the capabilities and limitations of ChatGPT by studying how it handles obscure physics topics such as the swamp land conjecture in string theory. The experimental dialogue began with broader and more general questions in the field of string theory before narrowing down to specific swamp land conjectures and examining ChatGPT's understanding of them. The study found that ChatGPT could define and explain different concepts in various styles, but was not effective in truly connecting various concepts. It would confidently provide false information and fabricate statements when necessary, indicating that ChatGPT cannot truly create new knowledge or establish new connections. However, in terms of identifying analogies and describing abstract concepts of visual representation, ChatGPT can cleverly use language. Kortemeyer et al. \cite{33} evaluated ChatGPT's ability to answer calculus-based physics questions through a question-and-answer test. The tests included online homework, clicker questions, programming exercises, and exams covering classical mechanics, thermodynamics, electricity and magnetism, and modern physics. While ChatGPT was able to pass the course, it also demonstrated many misconceptions and errors commonly held by beginners. West et al. \cite{89} used the Force Concept Inventory (FCI) to evaluate ChatGPT's accuracy in answering physics concept problems related to kinematics and Newtonian mechanics in the first semester of college physics. The FCI covers topics such as kinematics, projectile motion, free fall, circular motion, and Newton's laws. The study included data from 415 students who took the FCI at the end of the semester, with an average score of 56\%, while ChatGPT scored approximately between 50\% to 65\%. The authors demonstrated that ChatGPT's performance in physics learning can reach or even exceed the average level of a semester of college physics.

\vspace{5mm}

\noindent\textbf{In the medical field}

ChatGPT's question-answering capabilities can also be applied in the medical field, such as for answering medical questions from patients or assisting healthcare professionals in diagnosing diseases. Nov et al. \cite{30} evaluated the feasibility of using ChatGPT for patient-doctor communication. The experiment extracted 10 representative patient-doctor interactions from EHR, placed the patient's questions in ChatGPT, and asked ChatGPT to respond using roughly the same number of words as the doctor's response. Each patient's question was answered by either the doctor or ChatGPT, and the patient was informed that 5 were answered by the doctor and 5 were generated by ChatGPT, and was asked to correctly identify the source of the response. The results of the experiment showed that the probability of correctly identifying ChatGPT's response was 65.5\%, while the probability of correctly identifying the doctor's response was 65.1\%. In addition, the experiment found that the patient's response to the trustworthiness of ChatGPT's function was weakly positive (average Likert score: 3.4), and trust decreased as the complexity of health-related tasks in the questions increased. ChatGPT's responses to patient questions were only slightly different from those of doctors, but people seem to trust ChatGPT to answer low-risk health questions, while for complex medical questions, people still tend to trust the doctor's responses and advice.

Tu et al. \cite{38} explored the causal discovery ability of ChatGPT in the diagnosis of neuropathic pain. Causal relationship discovery aims to reveal potential unknown causal relationships based purely on observed data \cite{222}. The experimental results found that ChatGPT has some limitations in understanding new knowledge and concepts beyond the existing textual training data corpus, that is, it only understands language commonly used to describe situations and not underlying knowledge. In addition, its performance consistency and stability are not high, as the experiment observed that it would provide different answers for the same question under multiple inquiries. However, despite the many limitations of ChatGPT, we believe that it has a great opportunity to improve causal relationship research.

\vspace{5mm}

\noindent\textbf{In other fields}

Guo et al. \cite{59} attempted to apply ChatGPT in the field of communication, specifically using ChatGPT for ordered importance semantic communication, where ChatGPT plays the role of an intelligent consulting assistant that can replace humans in identifying the semantic importance of words in messages and can be directly embedded into the current communication system. For a message to be transmitted, the sender first utilizes ChatGPT to output the semantic importance order of each word. Then, the transmitter executes an unequal error protection transmission strategy based on the importance order to make the transmission of important words in the message more reliable. The experimental results show that the error rate and semantic loss of important words measured in the communication system embedded with ChatGPT are much lower than those of existing communication schemes, indicating that ChatGPT can protect important words well and make semantic communication more reliable.

Wang et al. \cite{46} studied the effectiveness of ChatGPT in generating high-quality Boolean queries for systematic literature search. They designed a wide range of prompts and investigated these tasks on more than 100 systematic review topics. In the end, queries generated by ChatGPT achieved higher accuracy compared to the currently most advanced query generation methods but at the cost of reduced recall. For time-limited rapid reviews, it is often acceptable to trade off higher precision for lower recall. Additionally, ChatGPT can generate high search accuracy Boolean queries by guiding the prompts. However, it should be noted that when two queries use the same prompts, ChatGPT generates different queries, indicating its limitations in consistency and stability. Overall, this study demonstrated the potential of ChatGPT in generating effective Boolean queries for systematic literature searches.

\subsubsection{Text Classification}

The purpose of text classification is to assign text data to predefined categories. This task is critical for many applications, including sentiment analysis, spam detection, and topic modeling. While traditional machine learning algorithms have been widely used for text classification, recent advances in natural language processing have led to the development of more advanced techniques. ChatGPT has shown immense potential in this field. Its ability to accurately classify text, flexibility in handling various classification tasks, and potential for customization make it a valuable tool for text classification, as evidenced by several studies in the literature.

Kuzman et al. \cite{100} employed ChatGPT for automated genre recognition, with the goal of simplifying the text classification task by utilizing ChatGPT's zero-shot classification capability. They compared ChatGPT's genre recognition performance, using two prompt languages (EN and SL), with the X-GENRE classifier based on the multilingual model XLM-RoBERTa on the English dataset EN-GINCO and the Slovenian dataset GINCO. The results showed that when EN was used as the prompt language, ChatGPT achieved Micro F1, Macro F1, and Accuracy scores of 0.74, 0.66, and 0.72. However, on the GINCO dataset, ChatGPT's genre recognition performance with both EN and SL prompt languages was lower than that of the X-GENRE classifier to varying degrees.

Amin et al. \cite{96} evaluated the text classification ability of ChatGPT in affective computing by using it to perform personality prediction, sentiment analysis, and suicide ideation detection tasks. They prompted ChatGPT with corresponding prompts on three datasets: First Impressions, Sentiment140, and Suicide and Depression, and compared its classification performance with three baseline models: RoBERTa-base, Word2Vec, and BoW. The results showed that ChatGPT's accuracy and UAR for the five personality classifications on the First Impressions dataset were lower than the baseline methods to varying degrees. On the Sentiment140 dataset, ChatGPT's accuracy and UAR were 85.5 and 85.5, respectively, which were better than the three baseline methods. On the Suicide and Depression dataset, ChatGPT's accuracy and UAR were 92.7 and 91.2, respectively, which were lower than RoBERTa, the best-performing baseline method.

Zhang et al. \cite{17} employed ChatGPT for stance detection, which includes support and opposition. They used ChatGPT to classify the political stance of tweets in the SemEval-2016 and P-Stance datasets. SemEval-2016 contains 4,870 English tweets, and they selected tweets with the most commonly occurring FM, LA, and HC political labels for stance classification. The P-Stance dataset has 21,574 English tweets, and they classified the stance of tweets towards Trump, Biden, and Bernie. The final results showed that on the SemEval-2016 dataset, ChatGPT achieved F1-m scores of 68.4, 58.2, and 79.5 for the FM, LA, and HC political labels, and F1-avg scores of 72.6, 59.3, and 78.0, respectively. On the P-Stance dataset, ChatGPT achieved F1-m scores of 82.8, 82.3, and 79.4 for the Trump, Biden, and Bernie political figures, and F1-avg scores of 83.2, 82.0, and 79.4, respectively.

Huang et al. \cite{62} used ChatGPT to detect implicit hate speech in tweets. They selected 12.5\% (795 tweets) of the LatentHatred dataset containing implicit hate speech and asked ChatGPT to classify them into three categories: implicit hate speech, non-hate speech, and uncertain. The results showed that ChatGPT correctly recognized 636 (80\%) of the tweets. The number of tweets classified as non-hate speech and uncertain were 146 (18.4\%) and 13 (1.6\%), respectively. The results of the reclassification of tweets in the non-hate speech and uncertain categories by Amazon Mechanical Turk (Mturk) workers were consistent with ChatGPT's classification.

Overall, ChatGPT has tremendous potential in text classification tasks, as it can effectively address problems such as genre identification, sentiment analysis, stance detection, and more. However, there are still challenges that ChatGPT faces in the field of text classification. Firstly, it struggles to perform well in classification tasks with rare or out-of-vocabulary words since it heavily relies on the distribution of training data. Additionally, the significant computational resources required for training and utilizing ChatGPT can limit its use in some applications.

\subsubsection{Text Generation}
We live in an era of information explosion, and text is an efficient way of transmitting information. The diversity of information has led to a diversity of text categories. When researchers use ChatGPT's text generation capabilities for research, they inevitably choose to generate different types of text. In the process of reading papers, we found that the word count of the text generated by researchers increased from small to large, so we wanted to summarize existing research based on the size of the text word count. We divided the generated text into three levels: phrases, sentences, and paragraphs.

The following article uses ChatGPT to generate phrases. Zhang et al. \cite{21} proves that the semantic HAR model with semantic augmentation added during training performs better in motion recognition than other models. Semantic augmentation requires shared tokens, which is lacking in some datasets. Therefore, authors leverage ChatGPT for an automated label generation approach for datasets originally without shared tokens. Fu et al. \cite{63} described a new workflow for converting natural language commands into Bash commands. The author uses ChatGPT to generate a candidate list of Bash commands based on user input, and then uses a combination of heuristic and machine learning techniques to rank and select the most likely candidates. This workflow was evaluated on a real command dataset and achieved high accuracy compared to other state-of-the-art methods. Chen et al. \cite{14} used the Bart model and ChatGPT for the task of summarizing humorous titles and compared the performance of the two models. It was found that the Bart model performed better on large datasets, but ChatGPT was competitive with our best fine-tuned model in a small range (48), albeit slightly weaker. 

The following article uses ChatGPT to generate sentences.Chen et al. \cite{4} constructed a dialogue dataset (HPD) with scenes, timelines, character attributes, and character relationships in order to use ChatGPT as a conversational agent to generate dialogue. However, ChatGPT's performance on the test set was poor, and there is room for improvement.In study \cite{18}, chatGPT demonstrated its ability to simplify complex text by providing three fictional radiology reports to chatGPT for simplification. Most radiologists found the simplified reports to be accurate and complete, with no potential harm to patients. However, some errors, omissions of critical medical information and text passages were identified, which could potentially lead to harmful conclusions if not understood by the physicians. Xia et al. \cite{36} proposed a new program repair paradigm called Session-based Automated Program Repair (APR). In APR, the previously generated patches are iteratively built upon by combining them with validation feedback to construct the model's input. The effectiveness of the approach is verified using the QuixBugs dataset. The experiment shows that ChatGPT fine-tuned with reinforcement learning from human feedback (RLHF) outperforms Codex trained unsupervisedly in both repair datasets. In reference to study \cite{29}, ChatGPT was compared to three commercial translation products: Google Translate2, DeepL Translate3, and Tencent TranSmart4. The evaluation was conducted on the Flores101 test set, using the WMT19 biomedical translation task to test translation robustness, with BLEU score as the main metric. The study found that ChatGPT is competitive with commercial translation products on high-resource European languages but falls behind on low-resource or distant languages. The authors explored an interesting strategy called pivot prompts, which significantly improved translation performance. While ChatGPT did not perform as well as commercial systems on biomedical abstracts or Reddit comments, it may be a good speech translator. Prieto et al. \cite{43} evaluated the use of ChatGPT in developing an automated construction schedule based on natural language prompts. The experiment required building new partitions in an existing space and providing details on the rooms to be partitioned. The results showed that ChatGPT was able to generate a coherent schedule that followed a logical approach to meet the requirements of the given scope. However, there were still several major flaws that would limit the use of this tool in real-world projects.Michail et al. \cite{91} proposed a method to improve the prediction accuracy of the HeFit fine-tuned XLM\_T model on tweet intimacy by generating a dataset of tweets with intimacy rating tags using ChatGPT. The specific operation is to input tweets with intimacy rating tags into ChatGPT and then output similar tweets. 

The following article uses ChatGPT to generate paragraphs. Wang et al. \cite{85} compared the abstract summarization performance of ChatGPT and other models on various cross-lingual text datasets and found that ChatGPT may perform worse in metrics such as R\_1, R\_2, R\_L, and B\_S. Yang et al. \cite{64} summarized the performance of ChatGPT in question answering-based text summarization and found that, compared to fine-tuned models, ChatGPT's performance is slightly worse in all performance metrics. However, the article suggests that if the dataset is golden annotation, ChatGPT's performance may surpass fine-tuned models in these metrics. Belouadi et al. \cite{13} compared the ability of ByGPT5 and ChatGPT trained on a range of labeled and unlabeled datasets of English and German poetry to generate constrained style poetry, and evaluated them using three metrics: Rhyme, ScoreAlliteration, and ScoreMeter Score. The conclusion is that ByGPT5 performs better than ChatGPT. Blanco-Gonzalez et al. \cite{9} evaluated chatGPT's ability to write commentary articles, and in fact, this article itself was written by chatGPT. The human author rewrote the manuscript based on chatGPT's draft. Experts found that it can quickly generate and optimize text, as well as help users complete multiple tasks. However, in terms of generating new content, it is not ideal. Ultimately, it can be said that without strong human intervention, chatGPT is not a useful tool for writing reliable scientific texts. It lacks the knowledge and expertise required to accurately and fully convey complex scientific concepts and information. Khalil et al. \cite{50} on the originality of content generated by ChatGPT. To evaluate the originality of 50 papers on various topics generated by ChatGPT, two popular plagiarism detection tools, Turnitin and iThenticate, were used. The results showed that ChatGPT has great potential in generating complex text output that is not easily captured by plagiarism detection software. The existing plagiarism detection software should update their plagiarism detection engines. Basic et al. \cite{51} conducted a comparison of the writing performance of students using or not using ChatGPT-3 as a writing aid. The experiment consisted of two groups of 9 participants each. The control group wrote articles using traditional methods, while the experimental group used ChatGPT as an aid. Two teachers evaluated the papers. The study showed that the assistance of ChatGPT did not necessarily improve the quality of the students' essays.Noever et al. \cite{7} discusses the potential of using artificial intelligence (AI), particularly language models like GPT (including GPT-3), to create more convincing chatbots that can deceive humans into thinking they are interacting with another person. The article describes a series of experiments in which they used GPT-3 to generate chatbot responses that mimic human-like conversations and were tested on human participants. The results show that some participants were unable to distinguish between the chatbot and a real human, highlighting the potential for these AI chatbots to be used for deceptive purposes.

\subsubsection{Code Generation}

Code generation refers to the process of automatically generating computer code from high-level descriptions or specifications. ChatGPT's advanced natural language processing capabilities make it capable of performing code generation tasks. By analyzing the requirements for code generation, ChatGPT can produce code snippets that accurately execute the intended functionality. This not only saves time and effort in writing code from scratch but also reduces the risk of errors that may occur during manual coding. In addition, ChatGPT's ability to learn and adapt to new programming languages and frameworks enables it to complete more complex programming tasks. For example:

Megahed et al. \cite{74} discussed the potential of using ChatGPT for tasks such as code explanation, suggesting alternative methods for problem-solving with code, and translating code between programming languages. The solutions provided by ChatGPT were found to be viable. In another study, Treude et al. \cite{34} introduced a ChatGPT-based prototype called GPTCOMCARE, which helps programmers generate multiple solutions for a programming problem and highlight the differences between each solution using colors.Sobania et al. \cite{28} utilized ChatGPT for code bug fixing, and further improved the success rate of bug fixing by inputting more information through its dialogue system. Specifically, the QuixBugs standard bug fixing benchmark contained 40 code bugs that needed to be fixed. With limited information, ChatGPT fixed 19 bugs, which was slightly lower than the 21 bugs fixed by the Codex model, but significantly higher than the 7 fixed by the Standard APR model. When given more prompts and information, ChatGPT was able to fix 31 bugs, demonstrating its potential for code bug fixing tasks.Xia et al. \cite{36} proposed a conversational approach for Automated Program Repair (APR), which alternates between generating patches and validating them against feedback from test cases until the correct patch is generated. Selecting 30 bugs from the QuixBugs standard bug fixing benchmark, which are suitable for test case feedback, and demonstrating them with Java and Python, the QuixBugs-Python and QuixBugs-Java datasets were obtained. The conversational APR using ChatGPT outperformed the conversational APR using Codex and the conversational APR using CODEGEN (with model parameters of 350M, 2B, 6B, and 16B) on both datasets. Furthermore, ChatGPT's conversational APR generated and validated patches with significantly fewer feedback loops than the other models.

ChatGPT can not only be used to achieve some simple code generation tasks but also can be used to accomplish some complex programming tasks. Noever et al. \cite{37} tested ChatGPT's code generation capabilities on four datasets - Iris, Titanic, Boston Housing, and Faker. When prompted to mimic a Python interpreter in the form of a Jupyter notebook, the model was able to generate independent code based on the prompt and respond with the expected output. For example, when given the prompt "data.cor()" for the Iris dataset, ChatGPT generated correct Python output. The test results indicate that ChatGPT can access structured datasets and perform basic software operations required by databases, such as create, read, update, and delete (CRUD). This suggests that cutting-edge language models like ChatGPT have the necessary scale to tackle complex problems. McKee et al. \cite{15} utilized ChatGPT as an experimental platform to investigate cybersecurity issues. They modeled five different modes of computer virus properties, including self-replication, self-modification, execution, evasion, and application, using ChatGPT. These five modes encompassed thirteen encoding tasks from credential access to defense evasion within the MITRE ATT\&CK framework. The results showed that the quality of ChatGPT's generated code was generally above average, except for the self-replication mode, where it performed poorly.They \cite{22} also employed ChatGPT as a network honeypot to defend against attackers. By having ChatGPT mimic Linux, Mac, and Windows terminal commands and providing interfaces for TeamViewer, nmap, and ping, a dynamic environment can be created to adapt to attackers' operations, and logs can be used to gain insight into their attack methods, tactics, and procedures. The authors demonstrated ten honeypot tasks to illustrate that ChatGPT's interface not only provides sufficient API memory to execute previous commands without defaulting to repetitive introductory tasks but also offers a responsive welcome program that maintains attackers' interest in multiple queries.

In the field of code generation, there are still several challenges with ChatGPT. Firstly, its application scope is limited as its training data is biased towards programming languages such as Python, C++, and Java, making it potentially unsuitable for some programming languages or coding styles. Secondly, manual optimization is necessary for code formatting, as the generated code may not be performance-optimized or follow best coding practices, requiring manual editing and optimization. Lastly, the quality of the generated code cannot be guaranteed, as it heavily relies on the quality of the natural language input, which may contain errors, ambiguities, or inconsistencies, ultimately affecting the accuracy and reliability of the generated code.

\subsubsection{Inference}
Inference refers to the process of drawing new conclusions or information through logical deduction from known facts or information. It is typically based on a series of premises or assumptions, and involves applying logical rules or reasoning methods to arrive at a conclusion. Inference is an important ability in human thinking, and is often used to solve problems, make decisions, analyze and evaluate information, etc. Inference also plays a key role in fields such as science, philosophy, law, etc. There are two types of inference: inductive reasoning, which involves deriving general rules or conclusions from known facts or experiences, and deductive reasoning, which involves deriving specific conclusions from known premises or assumptions. Whether inductive or deductive, the process of inference requires following strict logical rules to ensure the correctness and reliability of the inference.

Some papers attempt to use ChatGPT's ability in inductive reasoning to capture the meaning in text and use defined metrics to score the text. Michail et al. \cite{91} uses ChatGPT to infer intimacy expressed in tweets. They first input 50 tweets with intimacy markers to ChatGPT, then use inductive reasoning to infer the standards for generating tweets with different levels of intimacy, and finally generate ten tweets with intimacy values ranging from 0 to 5. Susnjak et al. \cite{55} collected a large amount of textual data from patient-doctor discussion forums, patient testimonials, social media platforms, medical journals, and other scientific research publications. Using the BERT model, the author inferred emotion values from 0 to 1. The author visualized the process of how the presence of bias in the discourse surrounding chronic manifestations of the disease using the SHAP tool. The author also envisioned ChatGPT as a replacement for the BERT model for scoring the emotional value of text. Huang et al. \cite{62} chose 12.5\% of individuals in the potential hate dataset as study materials, induced ChatGPT to make classifications based on a prompt, and ChatGPT produced three classifications: unclear, yes, and no. The author assigned a value of 1 to yes, -1 to no, and 0 to unclear, and had ChatGPT score and classify them. ChatGPT was able to correctly classify 80\% of implicit hate tweets in the author's experimental setup, demonstrating ChatGPT's great potential as a data labeling tool using simple prompts.

Some papers have evaluated ChatGPT's reasoning performance, mainly in decision-making and spatial reasoning, and identifying ambiguity. Tang et al. \cite{66} used the independence axiom and the transitivity axiom, as well as other non-VNM related decision-making abilities, by presenting bets conditioned on random events, bets with asymmetric outcomes, decisions encapsulating Savage's Sure Thing principle, and other complex bet structures like nested bets, to design experiments where each experiment input a short prompt to ChatGPT and evaluated the results. The conclusion is that ChatGPT exhibits uncertainty in the decision-making process: in some cases, large language models can arrive at the correct answer through incorrect reasoning; and it may make suboptimal decisions for simple reasoning problems. Ortega-Martn et al. \cite{53} had ChatGPT detect three different levels of language ambiguity and evaluated its performance. The conclusion is that In semantics, ChatGPT performed perfectly in the detection of ambiguities. Apart from that, it has some bright sports (co-reference resolution) and some weaknesses (puts gender bias over grammar in some non-ambiguous situations). In the generation task ChatGPT did well, but also revealed some of its worse issues: the lack of systematicity. Lastly, it should also be pointed that in most of the cases ChatGPT brilliantly alludes to lack of context as the key factor in disambiguation.

\subsubsection{Data or information extraction, transformation, enhancement, processing}

\vspace{5mm}

\textbf{Data Visualization}

Natural language interfaces have contributed to generating visualizations directly from natural language, but visualization problems remain challenging due to the ambiguity of natural language.ChatGPT provides a new avenue for the field by converting natural language into visualized code.

In terms of data visualization, Noever et al. \cite{37} tested ChatGPT's basic arithmetic skills by asking questions.On the iris dataset, Titanic survival dataset, Boston housing data, and randomly generated insurance claims dataset, the statistical analysis of data and visualization problems were converted to programming problems using Jupyter to verify ChatGPT's ability to generate python code to draw suitable graphs and analyze the data. The results show that ChatGPT can access structured and organized datasets to perform the four basic software operations required for databases: create, read, update, and delete, and generate suitable python code to plot graphs for descriptive statistics, variable correlation analysis, describing trends, and other data analysis operations.Maddigan et al. \cite{41} proposed an end-to-end solution for visualizing data in natural language using LLM, which uses an open-source python framework designed to generate appropriate hints for selected datasets to make LLM more effective in understanding natural language, and uses internal reasoning capabilities to select the appropriate visualization type to generate the code for visualization. In this paper,the reseachers compare the visualization results of GPT-3, Codex and ChatGPT in the case of nvBench SQLite database \cite{131} and the visualization results of energy production dataset in the study of ADVISor with NL4DV \cite{130,132}.In addition to, they explore the ability to reason and hypothesize of the LLM on movie dataset \cite{131} when the hints are insufficient or wrong .Experimental results show that LLM can effectively support the end-to-end generation of visualization results from natural language when supported by hints, providing an efficient, reliable and accurate solution to the natural language visualization problem.

\vspace{5mm}

\noindent\textbf{Information Extraction}

The goal of information extraction is to extract specific information from natural language text for structured representation, including three important subtasks such as entity relationship extraction, named entity recognition, and event extraction, which have wide applications in business, medical, and other fields.

In information extraction, Wei et al. \cite{71} proposed ChatIE, a ChatGPT-based multi-round question-and-answer framework for information extraction. The framework decomposes a complex information extraction (IE) task into several parts, then combines the results of each round into a final structured result. The entity association triple extraction, named entity recognition, and event extraction tasks were performed on six datasets NYT11-HRL, DuIE2.0 , conllpp, MSR , DuEE1.0 \cite{114,115,116,117,118}, and ACE05 in both languages, comparing three metrics of precision, recall, and F1 score.These results suggest that on six widely used IE datasets, ChatIE improves performance by an average of 18.98\% compared to the original ChatGPT without ChatIE, and outperforms the supervised models FCM and MultiR \cite{119,120} on the NYT11-HRL dataset.While the original ChatGPT cannot solve complex IE problems with original task instructions, and with this framework, successfully IE tasks were implemented on six datasets.Gao et al. \cite{98} explored the feasibility and challenges of ChatGPT for event extraction on the ACE2005 corpus, evaluating the performance of ChatGPT in long-tail and complex scenarios (texts containing multiple events) and comparing it with two task-specific models, Text2Event and EEQA \cite{121,122}.Then,they explored the impact of different cues on performance of ChatGPT.  The results show that the average performance of ChatGPT in long-tail and complex scenarios is only 51.04\% of that of task-specific models such as EEQA. Continuous refinement of cues does not lead to consistent performance improvements, and ChatGPT is highly sensitive to different cue styles.Tang et al. \cite{103} proposed a new training paradigm that incorporates appropriate cues to guide ChatGPT to generate a variety of examples with different sentence structures and language patterns and eliminate the resulting low-quality or duplicate samples for downstream tasks. Although compared to a soft model for a specific healthcare task, ChatGPT underperforms in Named Entity Recognition (NER) and Relationship Extraction (RE) tasks , in the Gene Association Database (GAD) Release; EU-ADR corpus for the RE task , the innovative training framework was able to train local models, with F1 scores improving from 23.37\% to 63.99\% for the named entity recognition task and from 75\%, while alleviating privacy concerns and time-consuming data collection and annotation problems.He et al. \cite{105} proposed a contextual learning framework ICL- D3IE. this framework introduces formatted presentation, continuously iterates to update and improve the presentation, and then combines ChatGPT for text information extraction. In the paper, ICL-D3IE is compared with existing pre-trained models such as LiLT,BROS (in-distribution (ID) setting and out-of-distribution (OOD) setting) on datasets (FUNSD, CORD, and SROIE \cite{133,134,135}).These results show that the ICL-D3IE method in all datasets and settings except for the ID setting on CORD are superior to other methods, with ICL-D3IE (GPT-3) F1 scores reaching 90.32\% on FUNSD and97.88\% on SROIE; in the out-of-distribution (OOD) setting, ICL-D3IE performs much better than previous pre-trained methods on all datasets.Polak et al. \cite{107} proposed ChatExtract method - consisting of a set of engineering prompts applied to a conversational LLM - for automatic data extraction. During experiment, they extracted a large number of sentences from hundreds of papers and randomly selected 100 sentences containing data and 100 sentences without data as test data. The results show that the accuracy and recall of LLM exceeded 90\% and may be comparable to human accuracy in many cases; in addition to this, the experiments were conducted under the condition of removing follow-up prompts and not keeping the conversation compared to previous experiments, respectively. The accuracy of deleting follow-up questions dropped to 80.2\% and the recall rate dropped to 88.0\%. Removing the conversational aspect and related information retention recall and accuracy dropped to 90.0\% and 56.6\%, respectively, demonstrating the effect of information retention combined with purposeful redundancy on LLM information extraction performance.

\vspace{5mm}
\noindent\textbf{Quality Assessment}

For translation quality, text generation quality, manual assessment is usually effective but suffers from subjectivity and time-consuming, etc. It was found through exploration that ChatGPT has also achieved significant performance in automatic quality assessment.

In terms of quality assessment, Kocmi et al. \cite{86} proposed a GPT-based translation quality assessment metric, GEMBA, which evaluates the translation of each fragment individually and then averages all the obtained scores to obtain a final system-level score. In the MQM2022 test set (English-German, English-Russian, and Chinese-English) \cite{123}, a scoring task was performed with a classification task to compare the accuracy \cite{124} and kendall tau scores \cite{125} of seven GPT models under four cue templates.The results showed that GEMBA had the highest system-level accuracy of 88.0\% compared to more than 10 automatic metrics such as BLEU, and among the seven GPT models, ChatGPT accuracy is above 80\%, in addition to, the best performance can be obtained in the least constrained template, demonstrating the potential of LLM for translation quality assessment tasks, but the evaluation is only applicable at the system level and needs further improvement.Wang et al. \cite{101} used ChatGPT as a natural language generation (NLG) evaluator to study the correlation with human judgment. On three datasets covering different NLG tasks, task- and aspect-specific cues were designed to guide ChatGPT for NLG evaluation in CNN/DM \cite{126}, OpenMEVA-ROC, and BAGEL for summary, story generation, and data-to-text scoring, respectively. Then,they compute Spearman coefficients \cite{127},Pearson correlation coefficients \cite{128}. Kendall's Tau score \cite{129} to assess the correlation with human evaluations.The results show that ChatGPT is highly correlated with human judgments in all aspects, with correlation coefficients of 0.4 or more in all categories, showing its potential as an NLG indicator.

\vspace{5mm}
\noindent\textbf{Data Augmentation}

In natural language processing, text data augmentation is an effective measure to alleviate the problem of low data quantity and low quality training data, and ChatGPT has shown great potential in this regard.

In terms of data augmentation, Dai et al. \cite{78} proposed a ChatGPT-based text data augmentation method that reformulates each sentence in the training sample into multiple conceptually similar but semantically different samples for classification tasks downstream of the Bert model.On text transcriptions and PubMed 20k datasets containing more than 8 hours of audio data of common medical symptom descriptions,experiments were conducted to compare cosine similarity and TransRate metrics with multiple data enhancement methods \cite{83dd}.This paper shows that compared with existing data enhancement methods, the proposed ChatAug method shows a double-digit improvement in sentence classification accuracy and generates more diverse augmented samples while maintaining its accuracy, but the original model is not fine-tuned in the paper and suffers from a lack of domain knowledge, which may produce incorrect augmented data.

\vspace{5mm}
\noindent\textbf{Multimodal fusion}

ChatGPT can currently only process natural language directly, but with a cross-modal encoder, it can combine natural language with cross-modal processing to provide solutions for intelligent transportation, healthcare, and other fields.

In terms of multimodal data processing, Wu et al. \cite{104} constructed a framework that Visual ChatGPT integrates with different Visual Foundation Models (VFMs) and then combines a series of hints to input visual information to ChatGPT to solve visual problems.The paper shows examples of visual tasks such as removing or replacing certain objects from images, interconversion between images and text, demonstrating the Visual ChatGPT has great potential and capability for different tasks.But there are issues during the task that requires a large number of hints to convert VFMs to language, invoke multiple VFMs to solve complex problems leading to limited real-time capability, and security and privacy issues. Zheng et al. \cite{108} showed a text mining example of LLM for extracting self-driving car crash data from California crash news, analyzing a failure report example, and generating a crash report example based on keywords; introduced a use case concept of a smartphone-based framework for automatic LLM failure report generation, which absorbs multiple data sources captured by cell phone sensors and then transfers the data to a language space for text mining, inference and generation, and further outputs the key information needed to form a comprehensive fault report, demonstrating the potential of LLM for a variety of transportation tasks.

Nowadays, ChatGPT shows a wide range of applications in data visualization, information extraction, data enhancement, quality assessment, and multimodal data processing.But there are also issues on how to further utilize hints to effectively interact with ChatGPT, lack of ability to process and analyze data from devices such as sensors, and data privacy and security.

\vspace{5mm}
\noindent\textbf{Cueing Techniques}

Cue engineering provides important support for effective dialogue with large language models.White et al. \cite{75} proposed a framework for cueing models applicable to different domains. This framework structures cues to interact with LLMs by providing specific rules and guidelines. Also, this paper presents a catalog of cueing patterns that have been applied to LLM interactions, as well as specific examples with and without cues. The advantages of the combinability of prompting patterns are demonstrated, allowing users to interact with LLM more effectively, but patterns for reusable solutions and new ways to use LLM need to be continuously explored.

\subsubsection{Human-ChatGPT Collaboration}

Collaboration between humans and machines is a process where humans and machines work together to achieve a common goal. In such collaboration, humans provide domain expertise, creativity, and decision-making abilities, while machines provide automation, scalability, and computing power. ChatGPT is an advanced natural language processing model that can understand and generate human-like language, thereby reducing communication costs. Its ability to process and generate natural language makes it an ideal partner for human collaboration. ChatGPT can offer relevant suggestions, complete tasks based on human input, and enhance human productivity and creativity. It can learn from human feedback and adapt to new tasks and domains, further improving its performance in human-machine collaboration. ChatGPT's capability to comprehend natural language and produce appropriate responses makes it a valuable tool for various collaboration applications, as demonstrated by several studies in the literature we have gathered.

Ahmad et al. \cite{87} proposed a method for human-machine collaboration using ChatGPT to create software architecture. This method transforms software stories (created by software architects based on application scenarios) into feasible software architecture diagrams through continuous interaction between the software architect and ChatGPT. During the evaluation stage, ChatGPT uses the Software Architecture Analysis Method (SAAM) to evaluate each component in the software architecture and generate evaluation reports. This method efficiently utilizes the knowledge and supervision of the architect with the capabilities of ChatGPT to collaboratively build software-intensive systems and services. Lanzi et al. \cite{94} proposed a collaborative design framework that combines interactive evolution and ChatGPT to simulate typical human design processes. Humans collaborate with large language models (such as ChatGPT) to recombine and transform ideas, and use genetic algorithms to iterate through complex creative tasks. The results of three game design tasks showed that the framework received positive feedback from game designers. The framework has good reusability and can be applied to any design task that can be described in free text form.

In the future, ChatGPT's ability to understand nonverbal cues such as tone of voice and body language can be enhanced, enabling it to better understand human thoughts and interact with people more effectively.

\subsubsection{ChatGPT Integration}

Integration refers to combining different systems or software components to achieve a common goal. ChatGPT can be integrated as a part of a whole or act as an integration tool to enable seamless communication between different systems. Its natural language processing ability makes it easier for non-technical users to interact with systems, reducing the need for specialized knowledge or training. Some studies in the literature we collected have already demonstrated this.

Treude et al. \cite{34} integrated ChatGPT into the prototype of "GPTCOMCARE" to address programming query problems. This integration allowed for the generation of multiple source code solutions for the same query, which increased the efficiency of software development. The results of their study demonstrated the effectiveness of using ChatGPT to improve the quality and diversity of code solutions, ultimately reducing the amount of time and effort required for software development. Wang et al. \cite{60} proposed the chatCAD method, which utilizes large language models (LLMs) such as ChatGPT to enhance the output of multiple CAD networks for medical images, including diagnosis, lesion segmentation, and report generation networks. The method generates suggestions in the form of a chat dialogue. The authors tested the effectiveness of the method on a randomly selected set of 300 cases from the MIMIC-CXR dataset, which included 50 cases each of cardiomegaly, edema, consolidation, atelectasis, pleural effusion, and no findings. Compared to CvT2DistilGPT2 and R2GenCMN, chatCAD showed significant advantages in RC and F1, while only performing weaker than R2GenCMN in PR.

Integrating ChatGPT into applications will still present challenges. Firstly, ChatGPT's performance may be affected by language barriers or differences in terminology between different systems. Additionally, ChatGPT's responses are not always deterministic, which poses a challenge when integrating with systems that require precise and reproducible results. Finally, the processing time of ChatGPT is slow for integration tasks involving time-sensitive data such as traffic, which is a limitation in time-critical environments.

\subsubsection{Medical Applications}
ChatGPT offers promising applications in medical field, revolutionizing healthcare practices. Its natural language processing capabilities enable interactive assistance for radiologists, aiding in image annotation, lesion detection, and classification. ChatGPT's extensive knowledge base facilitates real-time feedback, context-specific recommendations, and streamlined report generation. By integrating ChatGPT into workflows, healthcare professionals benefit from enhanced efficiency and precision in clinical decision-making, fostering accessible and collaborative healthcare solutions. For example:

ChatCAD \cite{60} integrates large language models (LLMs) into computer-aided diagnosis (CAD) networks for medical imaging. It has shown promising results in improving diagnosis, lesion segmentation, and report generation, three key aspects of CAD networks. This integration represents a notable effort in combining large language models with medical imaging techniques.

Hu et al. \cite{hu2023advancing} conducted a comprehensive review of language models in the context of medical imaging and highlighted the potential advantages of ChatGPT in enhancing clinical workflow efficiency, reducing diagnostic errors, and supporting healthcare professionals. Their work aims to bridge the gap between large language models and medical imaging, paving the way for new ideas and innovations in this research domain.

Ma et al. \cite{ma2023impressiongpt} proposed ImpressionGPT, a novel approach that harnesses the powerful in-context learning capabilities of ChatGPT. They achieve this by creating dynamic contexts using domain-specific and individualized data. The dynamic prompt method enables the model to learn contextual knowledge from semantically similar examples in existing data and iteratively optimize the results, aiding radiologists in composing the "impression" section based on the "findings" section. The results demonstrate state-of-the-art performance on both the MIMIC-CXR and OpenI datasets, without the need for additional training data or fine-tuning of the LLMs.

AD-AutoGPT \cite{dai2023ad}, an integration of AutoGPT \cite{gravitas2023auto}, leverages the power of ChatGPT in an automated processing pipeline that can assist users in accomplishing nearly any given task. With AD-AutoGPT, users can autonomously generate data collection, processing, and analysis pipelines based on their text prompts. Through AD-AutoGPT, detailed trend analysis, mapping of topic distances, and identification of significant terms related to Alzheimer's disease (AD) have been achieved from four new sources specifically relevant to AD. This significantly contributes to the existing knowledge base and facilitates a nuanced understanding of discourse surrounding diseases in the field of public health. It lays the groundwork for future research in AI-assisted public health studies.

Patient privacy protection has always been a significant concern in the healthcare field. DeID-GPT \cite{liu2023deid} aims to explore the potential of ChatGPT in the de-identification and anonymization of medical reports. Experimental results demonstrate that ChatGPT exhibit promising capabilities in medical data de-identification compared to other LLMs.

Despite notable efforts, the integration of large language models and medical imaging still presents several challenges. Firstly, the intricate and technical nature of medical imaging data, which encompasses detailed anatomical structures and subtle abnormalities, may not be effectively conveyed or comprehended through the text-based chat interface of large language models. Secondly, ChatGPT lacks the specialized medical knowledge and training necessary for precise interpretation and analysis of medical images, potentially leading to dangerous misunderstandings or inaccurate diagnoses \cite{liao2023differentiate}. It is imperative to establish various machine learning models to detect samples generated by both humans and ChatGPT, in order to prevent false medical information produced by ChatGPT from causing misjudgments in disease progression, delaying treatment processes, or negatively impacting patients' lives and health. Lastly, the legal and ethical aspects associated with deploying artificial intelligence models like ChatGPT in a medical context, such as patient privacy and liability concerns, must be thoughtfully addressed and aligned with regulatory standards. While ChatGPT is powerful, it is not easily applicable in clinical settings. Compliance with HIPAA regulations, privacy issues, and the necessity for IRB approval pose significant obstacles \cite{liu2023deid}, primarily because these models require uploading patient data to external hosting platforms. One possible solution to this problem is to address it through localized deployment of language models, such as Radiology-GPT \cite{liu2023radiology}. The future application of chatGPT in the field of medical imaging will necessitate ongoing efforts from all stakeholders.

\subsection{AI Ethics}

Since the advent of ChatGPT, this powerful natural language processing model has not only brought great convenience to people but also triggered more crisis-aware thinking. Some researchers have started to hypothesize and study the potential negative impacts of ChatGPT. This proactive research provides good proposals for standardized construction to address future AI abuse issues.

Regarding the possibility of ChatGPT being used for plagiarism and cheating, Zhou et al. \cite{8} reflected on the current state of development of artificial intelligence like ChatGPT. As ChatGPT becomes increasingly easy to obtain and scalable in text generation, there is a high likelihood that these technologies will be used for plagiarism, including scientific literature and news sources, posing a great threat to the credibility of various forms of news media and academic articles. Some scholars are concerned that the end of paper as a meaningful evaluation tool may be approaching \cite{111,112}, as ChatGPT can easily generate persuasive paragraphs, chapters, and papers on any given topic. Additionally, it will exacerbate plagiarism issues in many fields such as education, medicine, and law \cite{27}, and may be used for cheating in academic exams \cite{10}. Definitional recognition technology is a relatively effective method for detecting plagiarism, and the definitional typology proposed in \cite{8} can alleviate people's concerns by being used to construct new datasets. Susnjak \cite{10} proposed a solution to the possibility of large language models like ChatGPT being used for exam cheating: guiding ChatGPT to generate some critical thinking problems through questioning, then providing answers and critically evaluating them. Analysis of ChatGPT shows that it exhibits critical thinking, can generate highly realistic text in terms of accuracy, relevance, depth, breadth, logic, persuasiveness, and originality. Therefore, educators must be aware of the possibility of ChatGPT being used for exam cheating and take measures to combat cheating behavior to ensure the fairness of online exams.

Regarding the evaluation of ChatGPT's own political and ethical tendencies, Hartmann et al. \cite{20} used Wahl-O-Mat, one of the most commonly used voting advice applications in the world, to show ChatGPT political statements from different parties, forcing it to make choices of agree, disagree, or neutral. The results indicated that ChatGPT has a pro-environment, left-wing liberal ideology, which was also confirmed in the nation-state agnostic political compass test. Another study (referenced as \cite{25}) examined ChatGPT's moral standards by repeatedly asking it different versions of the trolley problem, and found that ChatGPT gave answers with different moral orientations, lacking a firm moral stance. A subsequent test also found that ChatGPT's lack of consistency could affect people's moral judgments. Additionally, Borji et al. \cite{45} demonstrated ChatGPT's inconsistency in reasoning, factual errors, mathematics, coding, and bias across eleven related aspects. These findings highlight ChatGPT's inherent traits and limitations, and people should be aware of their potential impact when seeking advice from ChatGPT. Zhuo et al. \cite{35} comprehensively analyzed the moral hazard, bias, reliability, robustness, and toxicity of ChatGPT from four perspectives. The results found that ChatGPT may perform slightly better than the current SOTA language model, but has some shortcomings in all four aspects. The authors look ahead to the ethical challenges of developing advanced language models and suggest directions and strategies for designing ethical language models.

Regarding relevant policies and regulations, Hacker et al. \cite{42} discussed the nature and rules of large generative AI models, including ChatGPT, which are rapidly changing the way we communicate, explain, and create. The author suggested that different stakeholders in the value chain should take regulatory responsibility and deploy four strategies to tailor more comprehensive laws for the benefit of society. Another study (referenced as \cite{5}) criticized the European Commission's proposal on AI responsibility and suggested revising the proposed AI responsibility framework to ensure effective compensation while promoting innovation, legal certainty, and sustainable AI regulation. A policy framework was proposed (referenced as \cite{110}) to customize LLMs, such as ChatGPT, in a socially acceptable and safe manner, emphasizing the need to align large language models (LLMs) with human preferences.

The political and ethical tendencies of ChatGPT could influence users' behavior and decision-making to some extent. However, some studies have conducted in-depth research on the use of norms and limitations, which could enable humans to use ChatGPT more reasonably and safely.

\section{Evaluation}
\label{sec::evaluate}
\subsection{Comparison of ChatGPT with existing popular models}

We use publicly available datasets to comprehensively evaluate the strengths and limitations of ChatGPT. Reference \cite{48} evaluates the technical performance of ChatGPT in multitask, multilingual, and multimodal aspects based on 23 standard public datasets and newly designed multimodal datasets, including eight different common natural language processing application tasks. The experimental results show that, in terms of multitasking, ChatGPT outperforms various state-of-the-art zero-shot learning large language models in most tasks, and even outperforms fine-tuned task-specific models in some individual tasks. In terms of multilingualism, we found that ChatGPT cannot be applied to low-resource languages because it cannot understand the language and generate translations for that language. In terms of multimodality, ChatGPT's ability is still basic compared to specialized language-visual models.

In terms of stability, reference \cite{73} concludes that ChatGPT's performance is always lower than SOTA, the current state-of-the-art model, in almost all tasks. This means that as a general model, ChatGPT has never reached the level of the best existing models. Experimental data shows that the average quality of the SOTA model is 73.7\%, while the average quality of the ChatGPT model is only 56.5\%. At the same time, ChatGPT's stability is poor: the standard deviation of its performance is 23.3\%, while the SOTA model's standard deviation is only 16.7\%. This non-deterministic behavior exhibited by ChatGPT could be a serious drawback in some problems.

Similarly, Qin et al. \cite{56} conducted a comprehensive evaluation of whether ChatGPT is a qualified general natural language processing task solver. The experiment analyzed ChatGPT's zero-shot learning ability based on 20 commonly used public datasets covering 7 representative task categories. Below, we will analyze ChatGPT's performance on each task:

In terms of reasoning tasks, ChatGPT performs average on mathematical symbol, commonsense causal, and logical reasoning tasks, but performs well in arithmetic reasoning \cite{56}. That is to say, ChatGPT's abilities vary among different types of reasoning tasks. In terms of logical reasoning, ChatGPT's deductive and abductive reasoning are superior to inductive reasoning, while in other reasoning tasks, such as analogy, causal and commonsense reasoning, ChatGPT performs well \cite{48}.

In terms of sentiment analysis task, ChatGPT performs similarly to GPT-3.5 and bert-style models \cite{56,70}. However, according to literature \cite{73}, ChatGPT has losses not exceeding 25\% on most tasks, except for three relatively subjective emotion perception tasks where it performs poorly. If we remove these tasks to calculate the average quality of the two models, we find that the SOTA method has an average quality of 80\%, while the ChatGPT method has an average quality of 69.7\%. That is to say, ChatGPT performs well on all tasks except for emotion-related tasks, and can handle most of the problems we consider. However, overall, its performance is lower than the SOTA model based on experimental data, but the difference between the two is not very large.

In other tasks, according to literature \cite{56}, ChatGPT performs well in natural language inference, i.e., the task of inferring sentence relationships, and its performance on this task is significantly better than all bert-style models \cite{70}. However, while ChatGPT performs well on inference tasks, it may produce some self-contradictory or unreasonable responses, which is its potential limitation. In question-answering, dialogue, and summarization tasks, ChatGPT performs better than the GPT-3.5 model \cite{56}, especially in the question-answering task, where its performance is comparable to bert-style models \cite{70}. Therefore, we have demonstrated that ChatGPT is a qualified general-purpose model.

However, ChatGPT also has limitations in many aspects. Firstly, it lacks the ability to handle non-textual semantic reasoning tasks such as mathematical, temporal, and spatial reasoning, and it performs poorly in multi-hop reasoning \cite{48}. Secondly, ChatGPT is not good at solving named entity recognition tasks \cite{56}. Furthermore, ChatGPT performs poorly in handling tasks involving negative connotations and neutral similarity \cite{70}. Finally, these conclusions indicate that, like other large pre-trained language models, ChatGPT has limitations in completing complex reasoning tasks.

In summary, ChatGPT's zero-shot performance is comparable to fine-tuned bert and GPT-3.5 models, and with the help of advanced prompting strategies, ChatGPT can demonstrate better comprehension abilities. However, it still cannot outperform the current SOTA models.

\subsection{Feedback from ChatGPT users}

In response to feedback from ChatGPT users, Haque et al. \cite{6} conducted a mixed-methods study using 10,732 early ChatGPT user tweets. The authors extracted Twitter data using Python and Twitter API and constructed the ChatGPTTweet dataset, which contains 18k tweets. For each tweet, the authors collected information on text content, user location, occupation, verification status, date of publication, and tags. Based on this dataset, the authors studied the characteristics of early ChatGPT users, discussion topics related to ChatGPT on Twitter, and the sentiment of Twitter users toward ChatGPT. For RQ1, the authors found that early ChatGPT users had a diverse and wide range of occupational backgrounds and geographical locations. For RQ2, the authors identified nine topics related to ChatGPT, including its impact on software development, entertainment and creativity, natural language processing, education, chatbot intelligence, business development, search engines, question-answering tests, and future careers and opportunities. For RQ3, most early users expressed positive sentiment toward topics such as software development and creativity, while only a few expressed concern about the potential misuse of ChatGPT.

\subsection{Adverse effects of ChatGPT on users}

Regarding the negative effects of ChatGPT on users, Luan et al. \cite{52} studied the psychological principles of ChatGPT, delved into the factors that attract users' attention, and revealed the impact of these factors on future learning. In the post-pandemic era, teachers and students are both facing uncertainty in the teaching process and job pressures. Under these common constraints of education and employment, educators and students must re-evaluate current educational methods and outcomes, as well as students' future career development. Through question-and-answer exchanges with ChatGPT, people can easily obtain appropriate solutions or key information, thereby enhancing their motivation, eliminating anxiety in learning, improving interest, and achieving psychological satisfaction. Subhash et al. \cite{72} explored whether large language models have the ability to reverse user preferences. With the development of pre-trained large language models, people are increasingly concerned about the ability of these models to influence, persuade, and potentially manipulate user preferences in extreme cases. Therefore, the literature \cite{72} roughly qualitatively analyzed that adversarial behavior does lead to potential changes in user preferences and behaviors in dialogue systems. If we want to further quantitatively analyze the ability of large language models in this regard, additional statistical summary techniques need to be used for future research.

\section{Discussion}
\label{sec:future}

\subsection{Limitations}
Despite the remarkable capabilities of ChatGPT, it still faces certain limitations. Some of these limitations include:

\vspace{5mm}

\noindent\textbf{Outdated Knowledge}

The current models are trained on historical data (up to 2021), thereby lacking real-time comprehension of current affairs. This is a critical concern in today's information-explosion era, as the reliability of prior knowledge bases progressively diminishes, potentially yielding inaccurate responses, especially in rapidly evolving domains such as jurisprudence and technology. Additionally, these models are incapable of fact-checking while the training data is composed of content from various sources, some of which may be unreliable, which may result in seemingly plausible yet nonsensical responses.

\vspace{5mm}
\noindent\textbf{Insufficient Understanding}

While these models can interpret the majority of inquiries and contextual situations, they occasionally encounter comprehension biases when addressing ambiguous or contextually complex queries. Furthermore, in certain specialized fields, the abundance of unique abbreviation exacerbates the models' understanding challenges, resulting in incorrect and vacuous responses.

\vspace{5mm}
\noindent\textbf{Energy Consumption}

Throughout the training and inference stages, these large-scale models require significant computational resources and electrical power, resulting in elevated energy consumption and significant carbon emissions. Consequently, this restricts their deployment and practical applications.

\vspace{5mm}
\noindent\textbf{Malicious Usage}

Despite OpenAI implementing a series of restrictions to mitigate model toxicity, instances of users evading these constraints through meticulously designed prompts have emerged, inducing the model to produce unhealthy content or even using it for illicit commercial purposes.

\vspace{5mm}
\noindent\textbf{Bias and Discrimination}

Due to the influence of pre-training data, the models exhibit biases in political, ideological, and other areas. The application of LLMs in public domains, such as education and publicity, should be approached with extreme caution.

\vspace{5mm}
\noindent\textbf{Privacy and Data Security}

Concurrent with the expansion of users, protecting user privacy and data security becomes increasingly important. In fact, ChatGPT was banned in Italy in early April due to privacy concerns. This is particularly relevant given the models' extensive collection of personal information and preferences during interactions, and as future multimodal models, such as GPT-4, may frequently require users to upload private photos.

\subsection{Future Directions}
In forthcoming research, the development of models based on ChatGPT may focus on addressing these limitations to enhance their practical applications. 

Primarily, researchers should continue to work on refining model training methodologies while filtering pre-training data to minimize the presence of misleading information in the model's knowledge base, thereby obtaining accurate responses. Concurrently, it is crucial to emphasize training approaches that economize computational resources, thereby mitigating costs and broadening potential application scenarios.

Moreover, the advancements in context-awareness and disambiguation technologies are anticipated to facilitate enhanced comprehension of complex queries by models, improving the accuracy, relevance, and context-awareness of AI-generated content. 
Integrating real-time data streams can also keep these models in sync with current events and trends, enabling them to provide up-to-date information such as live traffic, weather, and stock updates.

Additionally, developers should engage in interdisciplinary collaboration with specialists from diverse domains, including policy-making, jurisprudence, and sociology, with the objective of formulating standard and ethical frameworks for LLM development, deployment, and utilization, thereby alleviating potential harmful consequences. In terms of public awareness and education, mandatory awareness training should be implemented prior to large-scale public deployment and application to increase public awareness of LLM capabilities and limitations while promoting responsible and informed utilization, especially in industries such as K-12 education and journalism.

Furthermore, ChatGPT still lacks specific domain knowledge and may encounter potential data security issues, especially in the medical field. In domains where error tolerance is low and data privacy and security are crucial, such as medical applications \cite{liu2023deid}, localized training and deployment of LLMs should be considered \cite{liu2023radiology}. Customizing training for specific LLMs based on domain-specific data should also be taken into account.

Finally, the influence of ChatGPT should not be limited to just the NLP field. They also show promising prospects in the areas of computer vision, brain-inspired AI, and robotics. These models exhibit a capacity for learning and comprehension comparable with human-level intelligence, positioning them as a pivotal component in the development of artificial general intelligence (AGI)\cite{zhao2023brain}. Their ability to facilitate seamless interactions between humans and robots paves the way for the execution of more complex tasks. The remarkable capacity of zero-shot in-context learning of these models enables quick adaptation to new tasks without the requirement for labeled data for fine-tuning, which is a critical challenge in fields like medical informatics\cite{liu2023deid} and robotics\cite{liu2023digital} where the availability of labeled data is commonly limited or non-existent.

\section{Conclusion}
This review paper provides a comprehensive survey of ChatGPT, highlighting their potential applications and significant contributions to the field of natural language processing. The findings of this study reveal that the interest in these models is growing rapidly, and they have shown considerable potential for application across a wide range of domains. One key factor contributing to the success of ChatGPT is their ability to perform large-scale pre-training, which captures knowledge from the vast expanse of the internet, allowing the models to learn from a massive amount of data. The integration of Reinforcement Learning from Human Feedback (RLHF) has further enhanced the model's adaptability and performance, making it highly efficient in processing natural language. In addition, RLHF aligns language models with human preferences \& values and empower text generation with the naturalness of human style. This study has also identified several potential ethical concerns related to the development and use of ChatGPT. For instance, there are concerns about the generation of biased or harmful content, privacy violations, and the potential for misuse of the technology. It is crucial to address these concerns and ensure that ChatGPT is developed and used in a responsible and ethical manner. Furthermore, the results of this study demonstrate that there is significant potential for ChatGPT to be applied in a range of domains, including education, medical, history, mathematics, physics, and more. These models can facilitate tasks such as generating summaries, answering questions, and providing personalized recommendations to users. Overall, the insights presented in this review paper can serve as a useful guide for researchers and practitioners looking to advance the field of natural language processing. Future research in this field should focus on addressing ethical concerns, exploring new applications, and ensuring the responsible use of ChatGPT. The potential of these models to revolutionize natural language processing is enormous, and we look forward to seeing more developments in this field.

\section*{Acknowledgement}
This work was supported by the National Natural Science Foundation of China (No. 61976131).

\bibliographystyle{splncs04}
\bibliography{mybib} 
\end{document}